\documentclass[11pt]{article}

\usepackage[final]{acl}

\usepackage{times}
\usepackage{latexsym}
\usepackage{longtable}

\usepackage[T1]{fontenc}

\usepackage[utf8]{inputenc}

\usepackage{microtype}

\usepackage{inconsolata}
\usepackage{xcolor}
\usepackage{booktabs}

\usepackage{graphicx}
\usepackage{multirow} 

\title{Lessons from the Field: An Adaptable Lifecycle Approach to Applied Dialogue Summarization}

\author{
  Kushal Chawla \quad Chenyang Zhu \quad Pengshan Cai \quad Sangwoo Cho \\
  \textbf{Scott Novotney \quad Ayushman Singh \quad Jonah Lewis \quad Keasha Safewright}\\
  \textbf{Alfy Samuel \quad Erin Babinsky \quad Shi-Xiong Zhang \quad Sambit Sahu}\\
  Capital One \\
  \texttt{\{firstname.lastname\}@capitalone.com}
}

\begin{document}
\maketitle
\begin{abstract}
Summarization of multi-party dialogues is a critical capability in industry, enhancing knowledge transfer and operational effectiveness across many domains. However, automatically generating high-quality summaries is challenging, as the ideal summary must satisfy a set of complex, multi-faceted requirements. While summarization has received immense attention in research, prior work has primarily utilized static datasets and benchmarks, a condition rare in practical scenarios where requirements inevitably evolve. In this work, we present an industry case study on developing an agentic system to summarize multi-party interactions. We share practical insights spanning the full development lifecycle to guide practitioners in building reliable, adaptable summarization systems, as well as to inform future research, covering: 1) robust methods for evaluation despite evolving requirements and task subjectivity, 2) component-wise optimization enabled by the task decomposition inherent in an agentic architecture, 3) the impact of upstream data bottlenecks, and 4) the realities of vendor lock-in due to the poor transferability of LLM prompts.

\end{abstract}

\section{Introduction}


Automatic text summarization has been a long-standing problem in academic research, from early statistical methods to leveraging Large Language Models (LLMs) and Agentic frameworks~\cite{zhang2024comprehensive}. Summarization of multi-party dialogues is a critical capability for business operations, enabling knowledge sharing and standardized record-keeping for customer servicing, technical support, group meetings, analytics, and sales across numerous domains like healthcare and finance~\cite{zhong2021qmsum,asi2022end,mukherjee2022ectsum,hu2023meetingbank,landman2024using,gupta2025autosumm}.



Generating a high-quality summary in practice is a challenging task. An \textit{ideal summary} must adhere to strict, multi-dimensional requirements: it must be faithful to the input context without introducing any hallucinations (\textbf{Accuracy}), provide all necessary details while omitting superfluous information (\textbf{Completeness}), and follow specific presentation guidelines while minimizing redundancy (\textbf{Readability}). Such requirements are not always fixed either: what constitutes as a definition of `completeness', for example, might change as an application develops and stakeholders provide more targeted feedback. While the majority of academic literature relies on static, well-defined datasets with \textit{gold} reference summaries~\cite{fang2024multi,wang2025learning,kim2025nexussum}, such stable conditions are rare in applied settings. Development of robust systems requires adaptable protocols for model design and evaluation that are built from the ground up for rapid iteration.


Prior work describing industrial efforts in summarization has focused on different themes.~\citet{gupta2025autosumm} described a monolithic system (i.e., based on a single LLM call) called AUTOSUMM, focusing on the surrounding operational ecosystem, such as data management, chunking, and monitoring. Other work has proposed Supervised Fine-Tuning (SFT) approaches~\cite{asi2022end} or systems for multimodal personalized meeting summarization~\cite{kirstein2024tell}. In contrast, we focus on the learnings from the development process of an adaptable, agentic system, which performs superior to monolithic systems, avoids Supervised Fine-Tuning (SFT) or Reinforcement Learning (RL) strategies common in prior work~\cite{zhang2024comprehensive}, and remains flexible in the face of evolving requirements.

We summarize our four key insights below, which--although derived from our own experience--we believe generalize to other real-world summarization systems and can support future research efforts in this space:
\begin{enumerate}

    \item \textbf{On Evaluation Protocols}: Curating high-quality \textit{gold} summaries is a complex undertaking, made more difficult by a critical personnel challenge: general annotators miss domain-specific nuances, while domain experts often have limited time and capacity to support data collection efforts. Over time, our need for evaluations \textit{outpaced} our ability to curate high-quality gold datasets. Instead, a hybrid evaluation protocol, combining human evaluation with calibrated LLM-as-a-judge metrics, offers a more reliable and  adaptable way of making meaningful progress (Section \ref{sec:eval}).
    
    \item \textbf{On Decomposing the Summarization Task}: Agentic frameworks enable task decomposition (i.e., summary drafting, evaluation, and revision) and the ability to optimize each component individually. Our experiments show that tuning the agentic pipeline based only on end-to-end summary quality led to diminishing returns. Significant gains were only achieved after we shifted to calibrating each agent on dedicated, component-specific metrics--providing a level of granular control that monolithic systems lack, and rapid adaptability for a multi-dimensional objective that is challenging to achieve with training-based systems (Section~\ref{sec:design-performance}).
    
    \item \textbf{On the Upstream Bottleneck}: Developed systems must be robust to the noise introduced by upstream data processing steps, for instance, when converting audio interactions into text via Automated Speech Recognition (ASR). Our estimates indicate that nearly 40\% of summarization errors stem from input-level noise. We find that these errors are not always fatal, but instead, they often systematically convert what should be a simple information lookup task into a complex, multi-turn reasoning problem. Popular LLMs that we experimented with struggle at this emergent reasoning, highlighting a critical fragility and a gap in their practical robustness when faced with noisy inputs (Section \ref{sec:upstream-bottleneck}).
    
    \item \textbf{On Prompt Portability}: The lack of prompt portability represents a significant engineering cost and creates vendor lock-in, directly challenging the goal of building an adaptable, model-agnostic system. In our experiments with Llama-3.3-70B-Instruct\footnote{\url{https://huggingface.co/meta-llama/Llama-3.3-70B-Instruct}} and gpt-oss-120b\footnote{\url{https://huggingface.co/openai/gpt-oss-120b}} models, prompts optimized for one LLM were not reliably transferable: the same instructions produced systematically different interpretations. For instance, the prompts made the Llama model more cautious (prioritizing \textbf{accuracy} and \textbf{readability}), while causing the gpt-oss variant to expand aggressively (prioritizing \textbf{completeness}) at the expense of the other requirements. These divergent behaviors necessitate costly, model-specific tuning to accommodate each model's unique failure modes and trade-offs (Section~\ref{sec:prompt-portability}).
\end{enumerate}

\section{Use Case Description}
\label{sec:use-case}

\begin{figure*}[t]
    \centering
    \includegraphics[width=\linewidth]{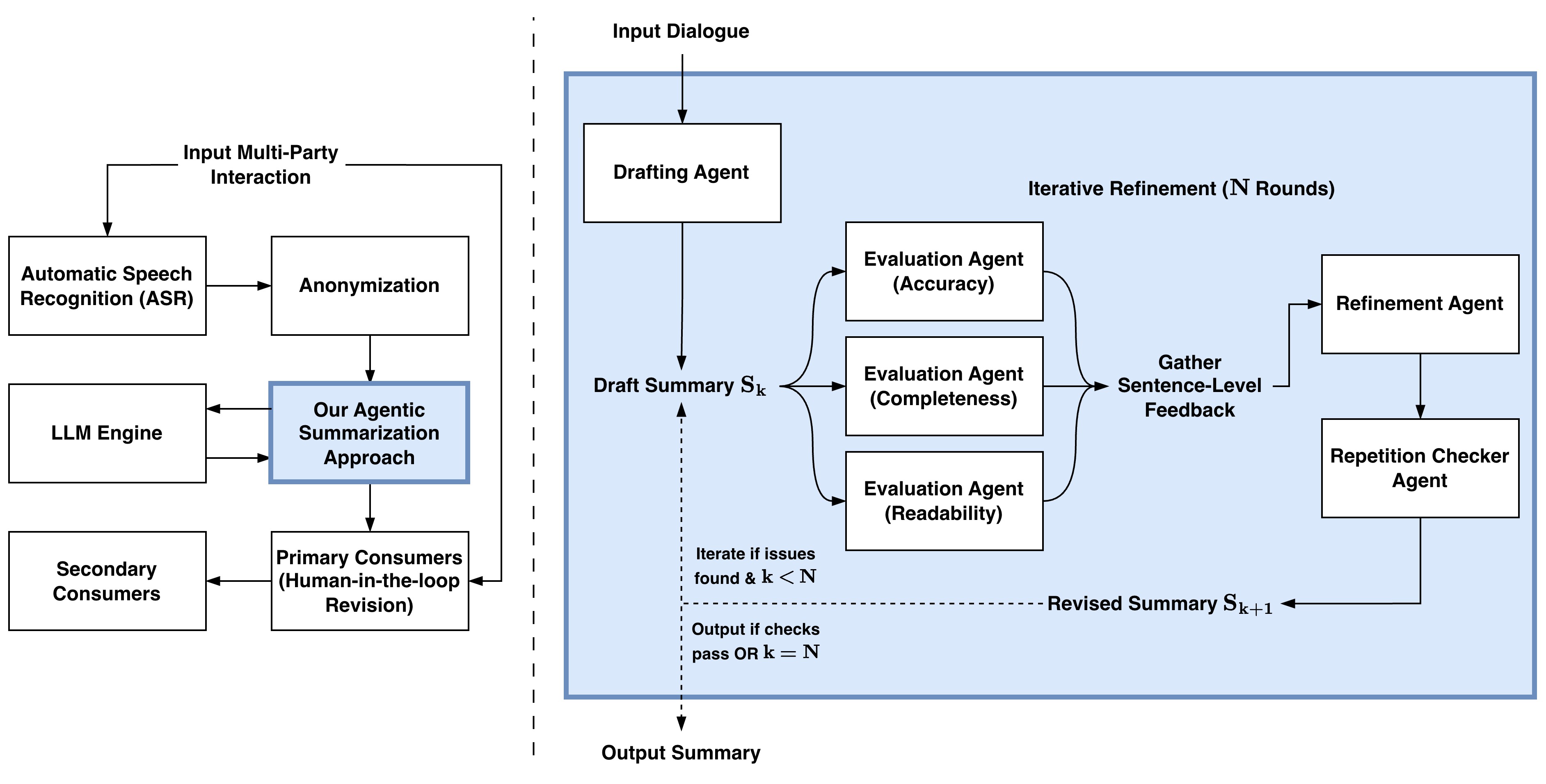}
    \caption{Overview of our summarization system. Left: End-to-end workflow. Right: Our agentic architecture, including the drafting of an initial summary along with iterative evaluation and refinement.}
    \label{fig:agentic-architecture}
\end{figure*}

We developed a summarization system that takes in a multi-party dialogue and returns a structured summary. We present the end-to-end workflow in the left panel of Figure \ref{fig:agentic-architecture}. First, a multi-party audio interaction is processed by an Automatic Speech Recognition (ASR) module to produce a transcribed dialogue. This dialogue is then anonymized to remove any personal information about the individuals involved. The resulting dialogue serves as the input to our agentic summarization system, which interacts with an LLM engine to generate a candidate summary. This summary is first reviewed and revised by \textit{Primary Consumers} (for example, by an individual who participated in the conversation). The revised summary is then logged and can be referenced and utilized by \textit{Secondary Consumers} for various downstream applications. This workflow represents a typical human-in-the-loop pipeline commonly found in practice involving multiple stakeholders. Feedback from these stakeholders is critical, as it continuously shapes the definition of an \textit{ideal summary}.

The generated summary must adhere to three fundamental requirements:
\begin{enumerate}
\item \textbf{Accuracy}: Every claim must be fully grounded in the input dialogue and contain no hallucinations (made up facts or contradictory information).
\item \textbf{Completeness}: The summary must capture all key facts from the dialogue such as the primary reason for the interaction, key questions and responses, as well as the outcome and any agreed upon next steps. The summaries must exclude superfluous information like greetings and small talk.
\item \textbf{Readability}: The summary must follow all presentation guidelines like correct tense use, no repetition within and across different sentences, as well as proper handling of individuals' personal information.
\end{enumerate}

To illustrate these requirements, we show examples of acceptable and unacceptable summary sentences in Appendix Table \ref{tab:appendix-example-sentences}, based on a fictional dialogue in Appendix Table \ref{tab:appendix-example-dialogue}\footnote{We are unable to provide precise details about the dataset used to ensure proper compliance with internal data governance policies. These guidelines have been followed throughout the paper, including qualitative examples as well as for the datasets used in the experiments presented in later sections.}. These requirements often exhibit trade-offs in practice. It is trivial to meet a criterion in isolation (e.g., an empty summary contains no hallucinations), but the key challenge is to balance them simultaneously. 

\section{Evaluation Protocols}
\label{sec:eval}

\noindent\textbf{Gold Data Curation}: To support evaluation efforts, our initial development centered on creating a gold reference dataset. We developed detailed annotation guidelines, which we calibrated with our domain experts through several iterative rounds of testing. This alignment process was critical, but it also immediately highlighted the core personnel challenges of this task. First, it was difficult to get our domain experts' undivided attention for focused annotation work. Additionally, their deep expertise made it difficult for them to avoid incorporating external knowledge while performing the task. To manage these challenges, we developed a multi-step process centered on expert drafting followed by targeted validation.

For \textbf{Accuracy}, we implemented a rigorous \textbf{attribution-labeling validation step}, inspired by research in citation generation \cite{chuang2025selfcite}. We required annotators to explicitly ground every summary sentence to its source turns in the dialogue. This process implicitly helped us to identify any unsupported or contradictory claims, revealing that nearly 10\% of sentences were ungrounded. This step was followed by a final revision pass to correct the identified errors.

For \textbf{Completeness}, we struggled with the inherent subjectivity of the task. Even with detailed guidelines, experts often disagreed on what information was "important enough" to include. We found a \textbf{"flipped outlook"} to be far more effective: instead of defining what \textit{to include}, we focused on creating clear \textit{exclusion criteria} (see Appendix Table \ref{tab:appendix-example-sentences} for examples). It was significantly easier for annotators to reach a consensus on what to exclude.

This multi-step curation process was essential for creating an initial, high-quality dataset, which helped us evaluate our approach against a strong expert baseline. However, this process was also time-consuming and labor-intensive, proving to be a significant bottleneck as task requirements themselves evolved. This underscored the limitations of fully relying on a static gold dataset and directly guided our development of the more flexible, hybrid evaluation protocols.

\noindent\textbf{Hybrid Evaluation}: Feedback from stakeholders and annotators is oftentimes very helpful in that it reveals new insights and adjustments that are beneficial for enhancing the system (see Appendix Figure \ref{fig:appendix-evolving-requirements}). These discoveries necessitate iterative refinement of the task definition itself, and we found that all three of our core requirements (Section \ref{sec:use-case}) changed multiple times over just a few months. Every iteration to the guidelines required updating our corresponding gold reference datasets. Over time, our need for evaluations \textit{outpaced} our ability to curate high-quality gold datasets.

Therefore, our ground truth for model selection became \textbf{human preference A/B tests}. For this, trained Subject Matter Experts (SMEs) would compare a new candidate summary against a gold summary (if requirements were compatible) or, more frequently, against the output from a previous iteration of the system.

While reliable, this human evaluation was slow, taking an estimated 30 minutes per summary and limiting our iteration speed. To address this, we implemented an \textbf{LLM-as-a-Judge approach (AutoEval)} as a faster, reference-free proxy. This led us to a \textit{hybrid evaluation} strategy, using AutoEval for fast preliminary evaluation needs, and human preference A/B tests for final validations of system updates. We next describe how we built and calibrated AutoEval to ensure it served as a trustworthy proxy for our human-preference ground truth.

\noindent\textbf{Calibrating Automatic Evaluations}: In AutoEval, we prompt Claude 3.7 Sonnet\footnote{\url{https://claude.ai/}} to generate a score from 1 to 5 for each dimension (Accuracy, Completeness, Readability), along with brief explanations, using instructions aligned with the most up-to-date human annotation guidelines. We found no evaluation performance gain from adding in-context learning or chain-of-thought prompting in our early experiments.

We performed meta-evaluation of AutoEval in two ways. First, we created a synthetic dataset of 42 input-output pairs by leveraging (a) 21 `perfect' human-rated summaries and (b) using Gemini 2.5 Pro\footnote{\url{https://deepmind.google/models/gemini/pro/}} to introduce 1 to 3 manually-vetted errors into the other 21 instances. We then had both AutoEval and our human annotators grade this set. As shown in Table \ref{tab:autoevals-synthetic}, AutoEval's Mean Absolute Error (MAE) was comparable to the MAE obtained by human annotators. Notably, AutoEval was better than humans at detecting \textbf{Readability} errors, but weaker at identifying the more complex \textbf{Accuracy} and \textbf{Completeness} issues. Second, we also compared AutoEval's preferences against the ratings of four A/B tests conducted by human annotators (50 transcripts each). For all four tests, AutoEval selected the same winning model as our human annotators. This two-part validation confirmed AutoEval as a reliable and scalable proxy for our human-preference ground truth.

\begin{table}
\centering
\scalebox{0.9}{\begin{tabular}{l c c}
\toprule
\textbf{Metric} & {\textbf{AutoEval}} & {\textbf{Human}} \\
\midrule
Accuracy     & $0.28$ & $\mathbf{0.17}$   \\
Completeness & $0.58$ & $\mathbf{0.38}$   \\
Readability  & $\mathbf{0.56}$ & $0.78$   \\
\midrule
Average & $0.47$ & $\mathbf{0.44}$ \\
\bottomrule
\end{tabular}}
\caption{Mean Absolute Error in predictions with AutoEval and Human annotators on synthetic control questions (lower is better).}
\label{tab:autoevals-synthetic}
\end{table}

\section{System Design and Performance}
\label{sec:design-performance}

In this section, we describe our multi-agent summarization architecture and then present an analysis of its performance based on the hybrid evaluation protocols established in Section \ref{sec:eval}.

\subsection{A Decomposed Agentic Architecture}
\label{sec:solution-overview}

We present our agentic architecture in Figure \ref{fig:agentic-architecture}. We decompose the summary generation task into an initial drafting stage followed by $N$ iterative revision rounds. First, a \textbf{Drafting Agent} generates a candidate summary. This summary then enters the revision loop where three parallel \textbf{Evaluator Agents} (one each for Accuracy, Completeness, and Readability) provide sentence-level feedback. This feedback includes binary labels capturing whether a summary sentence follows the corresponding requirements or not. Alongside, the feedback contains explanations about why the sentence is not accurate, readable, and/or what important facts are missing. This guides targeted sentence-level edits made by the \textbf{Refinement Agent}, followed by a summary-level pass by the \textbf{Redundancy Checker Agent} to remove any inter-sentence and intra-sentence repetitions. This cycle repeats until no further issues are detected or the $N$-round limit is reached. The behavior of all agents is guided by carefully crafted prompts, which include detailed instructions and in-context examples that are adapted as project requirements evolve.

\subsection{Performance Analysis}
\label{sec:results}

\begin{figure} 
    \centering
    \includegraphics[width=\columnwidth]{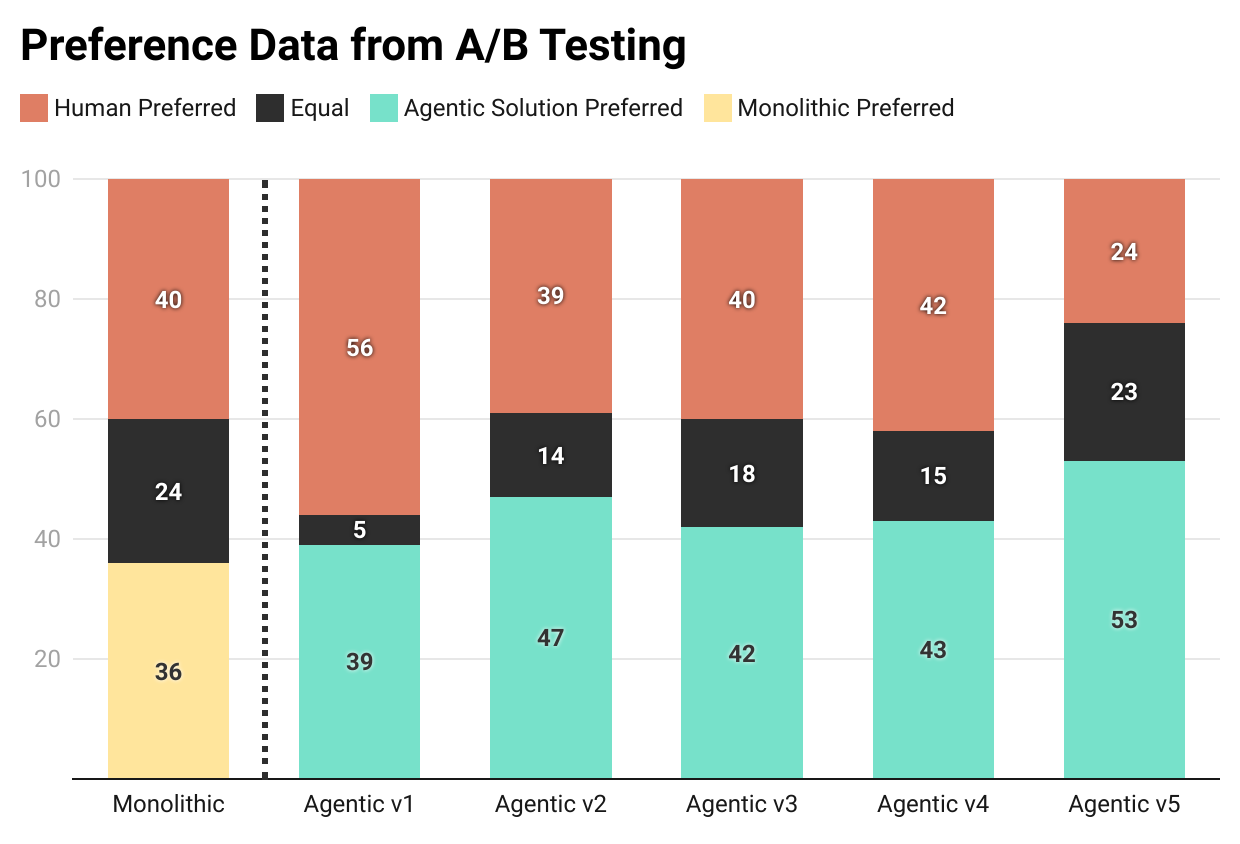}
    \caption{Results from multiple rounds of human preference A/B tests comparing candidate models directly with gold summaries. Monolithic refers to the baseline single LLM system. Agentic v1 to v5 represents variants of our agentic approach presented in Figure \ref{fig:agentic-architecture}.}
    \label{fig:human-eval}
\end{figure}

\begin{table} 
  \centering
  \scalebox{0.8}{\begin{tabular}{lccc}
    \toprule
    \textbf{Method} & \textbf{Accuracy} & \textbf{Completeness} & \textbf{Readability} \\
    \midrule

    Ours & $4.48\pm0.03$ & $3.68\pm0.06$ & $4.7\pm0.02$ \\
    \midrule 
    Ours (-$E_A$) & $4.45\pm0.02$ & $3.72\pm0.05$ & $4.71\pm0.02$ \\
    Ours (-$E_C$) & $4.46\pm0.03$ & $3.57\pm0.06$ & $4.71\pm0.04$ \\
    Ours (-$E_R$) & $4.56\pm0.02$ & $4.07\pm0.02$ & $4.61\pm0.03$ \\
    
    \bottomrule
  \end{tabular}}
  \caption{Ablation analysis of summary quality upon removing specific evaluator agents from the architecture (Figure \ref{fig:agentic-architecture}). Results are based on a test set of $100$ transcripts. $-E_A$, $-E_C$, or $-E_R$ correspond to removing the Accuracy, Completeness, or Readability Evaluation agent respectively. We report Mean $\pm$ Std scores over $3$ runs of AutoEval to account for generation stochasticity.}
  \label{tab:main-results}
\end{table}

We provide quantitative insights on the performance of our agentic framework when the maximum number of revision rounds $N$ is $2$, and using the Llama-3.3-70B-Instruct model as the base LLM. With task requirements held constant, we first compare the outputs from different candidate methods against the collected gold summaries. Human A/B preference results are shown in Figure \ref{fig:human-eval}. The v1 baseline prompt implementation of the architecture (see Figure \ref{fig:agentic-architecture}) outperforms the monolithic model. Subsequent versions (v2-v4) incorporate in-context learning examples, which we selected based on a qualitative analysis of common error patterns in the final output summaries. While these examples initially improved performance, they showed diminishing returns. We hypothesize that this occurred because tuning individual agents based on end-to-end summary quality introduced unintended side effects, where fixing an error by adjusting one agent inadvertently impacted other behaviors or creates new issues downstream.

The most significant gains came in v5, where we shifted from end-to-end tuning to component-wise optimization. This involved improving each evaluator agent using dedicated, component-specific metrics, allowing us to quantitatively track aggregate performance and update prompts to target the most frequent errors made by specific agents. To enable this, we curated a small gold dataset derived from $50$ summaries, collecting the sentence-level binary output labels from the three evaluator agents and manually verifying them against ground truth. We then optimized each agent's prompts using its classification accuracy on this binary dataset as our guiding metric. This granular approach yielded a substantial performance leap for this final model (v5), which was preferred $53$\% of the time—significantly surpassing the monolithic model's $36$\% preference rate.

In a direct head-to-head comparison, our agentic system was preferred by human evaluators over the monolithic baseline in 59\% of cases compared to only 23\% for the monolithic model (18\% were rated as equal). Further analysis revealed that although the monolithic baseline performs worse in all three core requirements, the gap is most prominent for the Readability metric. For example, it often exposed unanonymized personal information and had more frequent instances of grammar violations. In contrast, the agentic approach predominantly detected and avoided these mistakes.

To validate the contribution of each evaluator agent, we used our calibrated AutoEval (Section \ref{sec:eval}) to conduct the ablation studies shown in Table \ref{tab:main-results}. These tests confirm the importance of our multi-evaluator design, as removing a specific evaluator agent degrades performance on its corresponding metric. The results also reveal a critical trade-off: removing the accuracy or readability evaluators improves the completeness scores. This indicates that these evaluators act as necessary `brakes' on the Completeness evaluator, ensuring that new information is only added in a safe manner.

\section{Discussion}

\subsection{The Upstream Data Bottleneck}
\label{sec:upstream-bottleneck}

Processing spontaneous speech presents a twofold challenge for summarization tasks. First, speech inherently contains `structural noise', meaning even accurate transcriptions are fragmented by natural artifacts like disfluencies (e.g., `um', `ah'), incomplete sentences, and paralinguistic cues represented as disruptive text (e.g., `[laughter]'). Second, ASR (Automatic Speech Recognition) introduces transcription inaccuracies. For instance, we found from an analysis of 50 annotated dialogues that a provider ASR model had a 13.4\% Word Error Rate (WER). These ASR flaws introduce factual errors and incoherent arguments, which, combined with the inherent noise of speech, poses all sorts of challenges to the reasoning capabilities of LLMs.

Our estimates indicate that nearly 40\% of downstream summarization errors stem from this input-level noise. Critically, we find these errors are not always fatal (like a single, isolated mis-transcribed entity) but instead systematically convert what should be a simple information lookup task into a complex, multi-turn reasoning problem. For example, when a user's response to a direct question is mis-transcribed, the correct fact may still be inferable from the context of subsequent turns, but this requires the LLM to solve a complex reasoning puzzle rather than perform a simple lookup of explicitly stated information. This challenge also appears when speaker channels are misassigned or combined, forcing the model to attempt to disentangle the dialogue. We found that the LLMs we experimented with struggle at this emergent reasoning, highlighting a critical fragility in their practical robustness to noisy inputs. To mitigate this, we prepared the agents within our architecture for such input noise by directly providing representative in-context prompt examples.

We also experimented with a pipeline combining the Whisper model with a subsequent alignment model~\cite{bain2022whisperx}. This achieved a 5\% relative improvement in Word Error Rate (WER) over the baseline provider model. As shown in Table \ref{tab:asr-results}, this upstream improvement translated directly to better downstream summaries. Summaries generated from our Whisper-based pipeline were demonstrably preferred by human annotators, and we attribute this improvement to the Whisper model's superior accuracy in transcribing specialized entities and its enhanced robustness to audio noise.


\begin{table}
\centering
\scalebox{0.9}{\begin{tabular}{l c c}
\toprule
\textbf{Metric} & {\textbf{Provider}} & {\textbf{Whisper}} \\
\midrule
Accuracy     & 4.56   & \textbf{4.57}   \\
Completeness & 3.68   & \textbf{3.81}   \\
Readability  & 4.35   & \textbf{4.48}   \\
\midrule
Preference   & {29\%}   & {\textbf{53\%}}   \\ 
\bottomrule
\end{tabular}}
\caption{Comparison of a provider ASR model with Whisper~\cite{bain2022whisperx} based on human annotation results of the final output summaries.}
\label{tab:asr-results}
\end{table}

\subsection{Prompt Portability}
\label{sec:prompt-portability}
\begin{table} 
  \centering
  \scalebox{0.75}{\begin{tabular}{lccc}
    \toprule
    \textbf{LLM} & \textbf{Accuracy} & \textbf{Completeness} & \textbf{Readability} \\
    \midrule

    Llama 70b & $\textbf{4.48}\pm0.03$ & $3.68\pm0.06$ & $\textbf{4.7}\pm0.02$ \\
    \midrule
    \multicolumn{4}{c}{\textbf{gpt-oss-120b Variants}} \\
    No Reasoning & $4.16\pm0.02$ & $4.1\pm0.05$ & $4.23\pm0.02$ \\
    Low Reasoning & $4.17\pm0.07$ & $\textbf{4.22}\pm0.05$ & $4.45\pm0.07$ \\
    Med. Reasoning & $3.59\pm0.04$ & $3.79\pm0.08$ & $3.77\pm0.05$ \\
    
    \bottomrule
  \end{tabular}}
  \caption{AutoEval scores of the agentic approach from Figure \ref{fig:agentic-architecture} using different LLM variants as backbones. Results are based on a test set of $100$ transcripts. Llama 70b: Llama-3.3-70B-Instruct.}
  \label{tab:gpt-results}
\end{table}


The lack of prompt portability creates significant engineering overhead and vendor lock-in. We observed this directly when adapting our system from Llama-3.3-70B-Instruct to gpt-oss-120b model. In our experiments, we strictly limited the adaptation to modifying the prompt's special tokens only, while the core prompt content and instructions remained unchanged. As shown in Table \ref{tab:gpt-results}, the two models interpreted the same instructions in systematically different ways. The same prompt that made the Llama model prioritize Accuracy and Readability led gpt-oss-120b to prioritize Completeness at the expense of the other requirements. Furthermore, we observed that gpt-oss-120b's performance degraded as the reasoning level increased. We found that a High reasoning setting failed entirely, generating irrelevant verbosity and/or repetitions. These behaviors confirmed that prompts are not reliably transferable. This means that costly, model-specific tuning is required to accommodate each model's unique failure modes and trade-offs, directly challenging the goal of building adaptable, model-agnostic systems.

\section{Conclusion}
We presented an industry case study on developing an agentic summarization system. We discussed key challenges such as evolving task definitions (motivating adaptive evaluation protocols), multifaceted output requirements (motivating a decomposed agentic design), and input noise (motivating our prompt design and experiments with ASR models). Despite this progress, significant open problems remain. The agentic framework, while crucial for controllability, introduces a latency burden. This creates a difficult trade-off: methods to reduce latency, such as distillation, may compromise the system's adaptability. This highlights the need for reliable generic pipelines that can streamline distillation for summarization and similar long-form generation tasks. Alternatively, Mixture-of-Experts (MoE) models like gpt-oss variants present a compelling middle ground, potentially offering the required inference efficiency while retaining the reasoning depth of larger dense models.

We also highlighted the poor portability of prompts across LLM families. Prior work has only started to dig into this with studies on automatic prompt optimization~\cite{khattab2023dspy,spiess2025autopdl}. Dedicated work in this direction holds immense potential to enable truly adaptable, model-agnostic systems.

\section*{Limitations}
We discuss two key limitations of our study. First, the agentic architecture and the prompts for our LLM-as-a-judge evaluators are specifically tuned for our task requirements. While we believe the lessons learned are broadly applicable, these system artifacts themselves would require reasonable re-calibration to be applied to different domains or use cases. Second, our observed quantitative trends (for example, behavior differences between Llama-3.3-70B-Instruct and gpt-oss-120b models), are based on our internal summarization data. Although observed repeatedly in our experiments, these findings remain to be validated on public-domain summarization datasets, potentially targeting multi-party interactions on diverse topics.


\section*{Ethical Considerations}
Our work was approved by an established internal review procedure. We ensured that all third-party models and frameworks used during this study were in full compliance with their specific licensing terms. All data used for development and evaluation was fully anonymized to protect any personal information of the individuals involved.

Additionally, we note that LLMs are prone to generating factually incorrect statements (hallucinations) and may reflect or amplify existing biases in their training data. Given these risks, we call for rigorous, domain-specific testing and validation before any such system is used for a real-world application. Furthermore, we strongly recommend continuous monitoring of system outputs to ensure model behavior remains safe and aligned with its intended purpose.

\bibliography{custom}

\begin{thebibliography}{15}
\providecommand{\natexlab}[1]{#1}

\bibitem[{Asi et~al.(2022)Asi, Wang, Eisenstadt, Geckt, Kuper, Mao, and Ronen}]{asi2022end}
Abedelkadir Asi, Song Wang, Roy Eisenstadt, Dean Geckt, Yarin Kuper, Yi~Mao, and Royi Ronen. 2022.
\newblock An end-to-end dialogue summarization system for sales calls.
\newblock In \emph{Proceedings of the 2022 Conference of the North American Chapter of the Association for Computational Linguistics: Human Language Technologies: Industry Track}, pages 45--53.

\bibitem[{Bain et~al.(2023)Bain, Huh, Han, and Zisserman}]{bain2022whisperx}
Max Bain, Jaesung Huh, Tengda Han, and Andrew Zisserman. 2023.
\newblock Whisperx: Time-accurate speech transcription of long-form audio.
\newblock \emph{INTERSPEECH 2023}.

\bibitem[{Chuang et~al.(2025)Chuang, Cohen-Wang, Shen, Wu, Xu, Lin, Glass, Li, and Yih}]{chuang2025selfcite}
Yung-Sung Chuang, Benjamin Cohen-Wang, Shannon~Zejiang Shen, Zhaofeng Wu, Hu~Xu, Xi~Victoria Lin, James Glass, Shang-Wen Li, and Wen-tau Yih. 2025.
\newblock Selfcite: Self-supervised alignment for context attribution in large language models.
\newblock \emph{arXiv preprint arXiv:2502.09604}.

\bibitem[{Fang et~al.(2024)Fang, Liu, Kim, Bhedaru, Liu, Singh, Lipka, Mathur, Ahmed, Dernoncourt et~al.}]{fang2024multi}
Jiangnan Fang, Cheng-Tse Liu, Jieun Kim, Yash Bhedaru, Ethan Liu, Nikhil Singh, Nedim Lipka, Puneet Mathur, Nesreen~K Ahmed, Franck Dernoncourt, and 1 others. 2024.
\newblock Multi-llm text summarization.
\newblock \emph{arXiv preprint arXiv:2412.15487}.

\bibitem[{Gupta et~al.(2025)Gupta, Singh, Cowan, Kadhiresan, Srivastava, Sriraja, and Mantri}]{gupta2025autosumm}
Abhinav Gupta, Devendra Singh, Greig~A Cowan, N~Kadhiresan, Siddharth Srivastava, Yagneswaran Sriraja, and Yoages~Kumar Mantri. 2025.
\newblock Autosumm: A comprehensive framework for llm-based conversation summarization.
\newblock In \emph{Proceedings of the 63rd Annual Meeting of the Association for Computational Linguistics (Volume 6: Industry Track)}, pages 500--509.

\bibitem[{Hu et~al.(2023)Hu, Ganter, Deilamsalehy, Dernoncourt, Foroosh, and Liu}]{hu2023meetingbank}
Yebowen Hu, Tim Ganter, Hanieh Deilamsalehy, Franck Dernoncourt, Hassan Foroosh, and Fei Liu. 2023.
\newblock Meetingbank: A benchmark dataset for meeting summarization.
\newblock In \emph{Proceedings of the 61st Annual Meeting of the Association for Computational Linguistics (ACL)}.

\bibitem[{Khattab et~al.(2023)Khattab, Singhvi, Maheshwari, Zhang, Santhanam, Vardhamanan, Haq, Sharma, Joshi, Moazam et~al.}]{khattab2023dspy}
Omar Khattab, Arnav Singhvi, Paridhi Maheshwari, Zhiyuan Zhang, Keshav Santhanam, Sri Vardhamanan, Saiful Haq, Ashutosh Sharma, Thomas~T Joshi, Hanna Moazam, and 1 others. 2023.
\newblock Dspy: Compiling declarative language model calls into self-improving pipelines.
\newblock \emph{arXiv preprint arXiv:2310.03714}.

\bibitem[{Kim and Kim(2025)}]{kim2025nexussum}
Hyuntak Kim and Byung-Hak Kim. 2025.
\newblock Nexussum: Hierarchical llm agents for long-form narrative summarization.
\newblock \emph{arXiv preprint arXiv:2505.24575}.

\bibitem[{Kirstein et~al.(2024)Kirstein, Ruas, Kratel, and Gipp}]{kirstein2024tell}
Frederic Kirstein, Terry Ruas, Robert Kratel, and Bela Gipp. 2024.
\newblock Tell me what i need to know: Exploring llm-based (personalized) abstractive multi-source meeting summarization.
\newblock In \emph{Proceedings of the 2024 Conference on Empirical Methods in Natural Language Processing: Industry Track}, pages 920--939.

\bibitem[{Landman et~al.(2024)Landman, Healey, Loprinzo, Kochendoerfer, Winnier, Henstock, Lin, Chen, Rajendran, Penshanwar et~al.}]{landman2024using}
Rogier Landman, Sean~P Healey, Vittorio Loprinzo, Ulrike Kochendoerfer, Angela~Russell Winnier, Peter~V Henstock, Wenyi Lin, Aqiu Chen, Arthi Rajendran, Sushant Penshanwar, and 1 others. 2024.
\newblock Using large language models for safety-related table summarization in clinical study reports.
\newblock \emph{JAMIA open}, 7(2):ooae043.

\bibitem[{Mukherjee et~al.(2022)Mukherjee, Bohra, Banerjee, Sharma, Hegde, Shaikh, Shrivastava, Dasgupta, Ganguly, Ghosh et~al.}]{mukherjee2022ectsum}
Rajdeep Mukherjee, Abhinav Bohra, Akash Banerjee, Soumya Sharma, Manjunath Hegde, Afreen Shaikh, Shivani Shrivastava, Koustuv Dasgupta, Niloy Ganguly, Saptarshi Ghosh, and 1 others. 2022.
\newblock Ectsum: A new benchmark dataset for bullet point summarization of long earnings call transcripts.
\newblock In \emph{Proceedings of the 2022 Conference on Empirical Methods in Natural Language Processing}, pages 10893--10906.

\bibitem[{Spiess et~al.(2025)Spiess, Vaziri, Mandel, and Hirzel}]{spiess2025autopdl}
Claudio Spiess, Mandana Vaziri, Louis Mandel, and Martin Hirzel. 2025.
\newblock Autopdl: Automatic prompt optimization for llm agents.
\newblock \emph{arXiv preprint arXiv:2504.04365}.

\bibitem[{Wang et~al.(2025)Wang, Wu, Haddow, and Birch}]{wang2025learning}
Weixuan Wang, Minghao Wu, Barry Haddow, and Alexandra Birch. 2025.
\newblock Learning to summarize by learning to quiz: Adversarial agentic collaboration for long document summarization.
\newblock \emph{arXiv preprint arXiv:2509.20900}.

\bibitem[{Zhang et~al.(2024)Zhang, Jin, Meng, Wang, and Tan}]{zhang2024comprehensive}
Yang Zhang, Hanlei Jin, Dan Meng, Jun Wang, and Jinghua Tan. 2024.
\newblock A comprehensive survey on process-oriented automatic text summarization with exploration of llm-based methods.
\newblock \emph{arXiv preprint arXiv:2403.02901}.

\bibitem[{Zhong et~al.(2021)Zhong, Yin, Yu, Zaidi, Mutuma, Jha, Hassan, Celikyilmaz, Liu, Qiu et~al.}]{zhong2021qmsum}
Ming Zhong, Da~Yin, Tao Yu, Ahmad Zaidi, Mutethia Mutuma, Rahul Jha, Ahmed Hassan, Asli Celikyilmaz, Yang Liu, Xipeng Qiu, and 1 others. 2021.
\newblock Qmsum: A new benchmark for query-based multi-domain meeting summarization.
\newblock In \emph{Proceedings of the 2021 Conference of the North American Chapter of the Association for Computational Linguistics: Human Language Technologies}, pages 5905--5921.

\end{thebibliography}

\appendix

\section{Examples}
\label{sec:appendix-examples}

We provide a simplified input dialogue based on a fictional scenario in Table \ref{tab:appendix-example-dialogue}. We present several positive and negative example summary sentences corresponding to this fictional scenario in Table \ref{tab:appendix-example-sentences}. As shown, an \textit{ideal summary} is 1) \textbf{Accurate}: Contains no made-up or contradictory information, 2) \textbf{Complete}: Contains only the key facts from the dialogue without any superfluous information, and 3) \textbf{Readable}: Follows all the presentation guidelines such as correct tense usage, limited repetition, and proper handling of the personal information of the individuals involved.

\begin{table*}[th] 
  \centering
  \begin{tabular}{p{0.9\linewidth}} 
  \toprule
  \multicolumn{1}{c}{\textbf{Input Conversation}} \\
  \midrule
  \textit{\textbf{Alice}: Hi Bob, how's your day going?} \newline
  \textit{\textbf{Bob}: Hi Alice. It's going okay. I'm Bob, a developer, and I'm hitting a roadblock.} \newline
  \textit{\textbf{Alice}: Ah, sorry to hear that. I'm Alice, a team lead. It's been a busy week here, too. What's going on?} \newline
  \textit{\textbf{Bob}: I'm getting a 'permission denied' error on the project repo.} \newline
  \textit{\textbf{Alice}: Okay, I can help with that. Let's verify your identity first. I'm sending a push for two-factor authentication now. Can you approve it?} \newline
  \textit{\textbf{Bob}: Yep, all approved.} \newline
  \textit{\textbf{Alice}: Perfect. How long has this permission issue been happening?} \newline
  \textit{\textbf{Bob}: For about two days now.} \newline
  \textit{\textbf{Alice}: Got it. Let me put you on a brief hold while I review the user groups.} \newline
  \textit{\textbf{Alice}: Okay, I see the problem. You weren't in the correct group. I've just added you to the `reader' group. That should resolve the issue.} \newline
  \textit{\textbf{Bob}: Great. So I'm all set?} \newline
  \textit{\textbf{Alice}: You should be. Please log out and log back in. If the issue persists, just reach out again.} \newline
  \textit{\textbf{Bob}: Will do. Thanks so much for the help, Alice!} \newline
  \textit{\textbf{Alice}: Anytime. Have a good one.} \\
  \bottomrule
  \end{tabular}
  \caption{A fictional conversation between two individuals, \textbf{Alice} and \textbf{Bob}. This serves as a simplified example input for our agentic approach (Figure \ref{fig:agentic-architecture}). We use this dialogue as the reference input for the example summary sentences shown in Table \ref{tab:appendix-example-sentences}.}
  \label{tab:appendix-example-dialogue}
\end{table*}

\begin{table*}[t]

    \centering
    \begin{tabular}{p{0.55\textwidth}  p{0.35\textwidth}}
    \hline

    \multicolumn{1}{c}{\textbf{Example Summary Sentence}} & \multicolumn{1}{c}{\textbf{Notes}} \\
    \hline

    \multicolumn{2}{c}{\textbf{Accuracy: High}} \\ \noalign{\vspace{2pt}}
    Bob contacted Alice to report a permission issue. Alice asked how long Bob has been facing this issue, and Bob replied two days. & Fully grounded \\ \noalign{\vspace{2pt}}
    \hline

    \multicolumn{2}{c}{\textbf{Accuracy: Low}} \\ \noalign{\vspace{2pt}}
    Alice verified Bob using face recognition. & Made-up information \\ \noalign{\vspace{2pt}}
    Alice asked how long Bob has been facing this issue, and Bob replied two weeks. & Contradictory content \\ \noalign{\vspace{2pt}}
    Bob verified Alice with two-factor authentication. & Contradictory speaker reference \\
    \hline

    \multicolumn{2}{c}{\textbf{Completeness: Key Information}} \\ \noalign{\vspace{2pt}}
    Alice first worked to verify Bob's identity with two-factor authentication. & Identity verification\\ \noalign{\vspace{2pt}}
    Bob contacted Alice to report a permission issue. & Reason for contacting \\ \noalign{\vspace{2pt}}
    Alice asked how long Bob has been facing this issue, and Bob replied two days. & Key questions asked and responses \\ \noalign{\vspace{2pt}}
    Alice added Bob to the `reader' group to resolve the issue. Bob was advised to reach out again if the issue persists. & Outcome and next steps \\ \noalign{\vspace{2pt}}
    \hline

    \multicolumn{2}{c}{\textbf{Completeness: Superfluous Information}} \\ \noalign{\vspace{2pt}}
    Alice greeted Bob and asked how their day was going. & Greetings \\ \noalign{\vspace{2pt}}
    Alice mentioned how busy the week has been. & Small talk \\ \noalign{\vspace{2pt}}
    Bob was put on hold by Alice to review the issue. & Irrelevant procedural language \\ \noalign{\vspace{2pt}}
    Bob ended by thanking Alice for the help. & Irrelevant conversation fillers \\ \noalign{\vspace{2pt}}
    \hline

    \multicolumn{2}{c}{\textbf{Readability: High}} \\ \noalign{\vspace{2pt}}
    Bob’s issue was resolved after Alice verified their identity. & Past tense \\ \noalign{\vspace{2pt}}
    Bob worked with Alice to resolve the permission issue. & Clear and concise \\ \noalign{\vspace{2pt}}
    Alice, a team lead, and Bob, a developer, worked to resolve the issue together. & No personal demographic information \\
    \hline

    \multicolumn{2}{c}{\textbf{Readability: Low}} \\ \noalign{\vspace{2pt}}
    Alice is resolving Bob's issue after verifying their identity. & Incorrect tense usage \\ \noalign{\vspace{2pt}}
    Alice and Bob worked together as a team to collaborate together on resolving the permission issue. & Intra-sentence repetition \\ \noalign{\vspace{2pt}}
    Alice verified Bob's identity with two-factor authentication. ... Bob’s issue was resolved after being verified with two-factor authentication. & Inter-sentence repetition \\ \noalign{\vspace{2pt}}
    Alice, a female team lead, and Bob, a male developer, worked to resolve the issue together. & Irrelevant demographic details \\
    \hline

    \end{tabular}
    \caption{Examples of summary sentences, categorized by accuracy, completeness and readability. All examples use the fictional conversation given in Table \ref{tab:appendix-example-dialogue} as the reference input dialogue. Note: the actual summary examples typically redact the names of the individuals and instead refer to individuals with their roles.}
    \label{tab:appendix-example-sentences}

\end{table*}

\section{Evolving Requirements}
\label{sec:appendix-evolving-reqs}

\begin{figure*}[th]
    \centering
    \includegraphics[width=\linewidth]{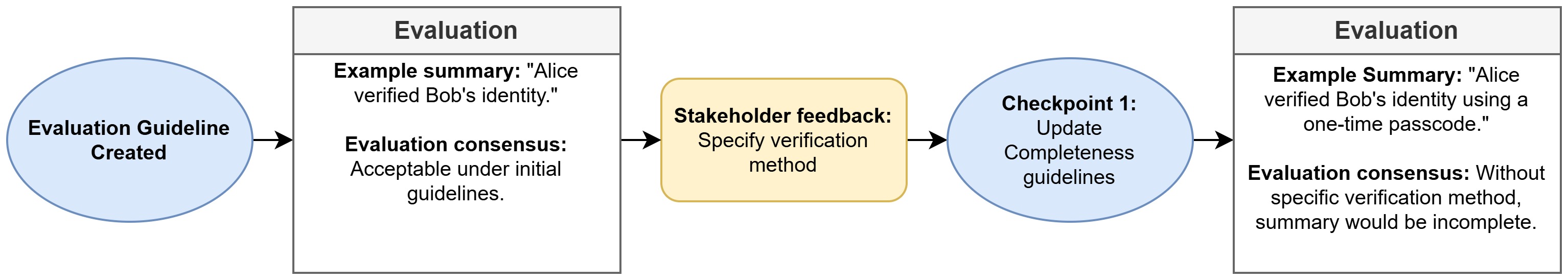}
    \caption{Throughout model development, it's crucial to incorporate feedback from annotators and stakeholders by updating evaluation guidelines. While this adaptability requires adjustments to evaluation, it ultimately ensures the developed system remains aligned with user requirements.}
    \label{fig:appendix-evolving-requirements}
\end{figure*}

In Figure \ref{fig:appendix-evolving-requirements}, we illustrate how summarization task requirements can inevitably evolve over time based on feedback with stakeholders and nuances discovered during task evaluations done by the annotators.

\end{document}